\documentclass[a4paper]{article}

\usepackage{hyperref}
\usepackage{INTERSPEECH2022}
\usepackage{amsmath}
\usepackage{booktabs}
\usepackage{caption}
\usepackage[autostyle, english = american]{csquotes}
\usepackage[inline]{enumitem}
\usepackage{multicol}
\usepackage{multirow}
\usepackage{numprint}
\usepackage{subcaption}
\usepackage{xcolor}
\MakeOuterQuote{"}

\usepackage[all=normal, paragraphs=tight, floats=tight, mathspacing=tight, leading=tight]{savetrees}
\everypar{\looseness=-1}
\npdecimalsign{.}
\npthousandsep{\,}

\makeatletter
\newcommand{\speciallabel}[2]{%
  \edef\@currentlabel{#1}\label{#2}%
}
\makeatother

\makeatletter
\newcommand{\smallsym}[2]{#1{\mathpalette\make@small@sym{#2}}}
\newcommand{\make@small@sym}[2]{%
  \vcenter{\hbox{$\m@th\downgrade@style#1#2$}}%
}
\newcommand{\downgrade@style}[1]{%
  \ifx#1\displaystyle\scriptstyle\else
    \ifx#1\textstyle\scriptstyle\else
      \scriptscriptstyle
  \fi\fi
}
\makeatother

\newcommand{\smallsim}{\smallsym{\mathrel}{\sim}}
\newcommand{\emailUrl}[1]{\href{mailto:#1@apple.com}{#1}}

\title{Space-Efficient Representation of Entity-centric Query Language Models\thanks{\hspace*{-1.8em}The template/entity lists used in this paper are available at\\ \href{https://github.com/apple/ml-interspeech2022-phi_rtn}{\texttt{github.com/apple/ml-interspeech2022-phi\_rtn}}.}}
\name{Christophe Van Gysel, Mirko Hannemann, Ernest Pusateri, Youssef Oualil, Ilya Oparin}
\address{Apple}
\email{\{\emailUrl{cvangysel}, \emailUrl{mirko\_hannemann}, \emailUrl{epusateri}, \emailUrl{youalil}, \emailUrl{ioparin}\}@apple.com}

\begin{document}

\newcommand{\LongMethodName}{$\phi\text{-RTN}$}
\newcommand{\Apply}[2]{#1\left(#2\right)}
\newcommand{\Prob}[1]{\Apply{P}{#1}}
\newcommand{\CondProb}[2]{\Prob{{#1} \mid {#2}}}
\newcommand{\BigOh}[1]{\mathcal{O}{(#1)}}
\newcommand{\compacttexttt}[1]{{\footnotesize \texttt{#1}}}
\newcommand{\SetSize}[1]{\left|#1\right|}

\maketitle
\begin{abstract}
Virtual assistants make use of automatic speech recognition (ASR) to help users answer entity-centric queries. However, spoken entity recognition is a difficult problem, due to the large number of frequently-changing named entities. In addition, resources available for recognition are constrained when ASR is performed on-device.
In this work, we investigate the use of probabilistic grammars as language models within the finite-state transducer (FST) framework. We introduce a deterministic approximation to probabilistic grammars that avoids the explicit expansion of non-terminals at model creation time, integrates directly with the FST framework, and is complementary to n-gram models.
We obtain a 10\% relative word error rate improvement on long tail entity queries compared to when a similarly-sized n-gram model is used without our method.
\end{abstract}

\section{Introduction}
\label{section:introduction}

Virtual assistants (VAs) are becoming increasingly popular \cite{Juniper2019popularity} to help users accomplish a variety of tasks \cite{Maarek2019alexa}. VA queries can be informational \cite{Broder2002taxonomy}, such as \emph{"how old is Joe Biden"}, or navigational/transactional (e.g., \emph{"show me Dune"}), and are often centered around a named entity. Spoken entity recognition is a difficult problem \cite{VanGysel2020entity,Gondala2021error,Saebi2021discriminative,Hayashi2020latent,Velikovich20128semanticlattice}, since the space of named entities is large, dynamic, and suffers from a long tail. In many use-cases, entity-oriented queries follow templates where the same template is applicable to entities of the same class \cite{Brown1992classlm,Shaik2012hierarchical}. Consider the previous example query \emph{"how old is Joe Biden"}, where segment \emph{"how old is"} references a specific entity attribute and \emph{"Joe Biden"} is an entity that belongs to the class of "famous people". We can now generalize this example into a probabilistic grammar that consists of the sole template \emph{"how old is X"} and $X$ is a non-terminal that expands to a weighted list of entity names. An additional benefit of this approach is that entities can be extracted from external knowledge sources \cite{VanGysel2020entity} that correlate with user queries and allow the VA to accurately recognize emerging entities that are not yet prevalent in training data.

In this paper we investigate the use of entity-centric probabilistic grammars as language models (LMs) within Automated Speech Recognition (ASR) to improve the recognition of entity-oriented queries that are common in VA usage. We are particularly interested in increasing the ASR system's coverage of entities, while minimizing model size---i.e., we wish to reduce the word error-rate of all entity-oriented queries represented by the grammar. Hence, methods are trained using the full grammar and evaluated on their ability to cover that same grammar.

Using grammars for ASR is not a new idea. Stolcke et al. \cite{Stolcke1994ngramscfg} presented an algorithm to estimate n-gram probabilities directly from rule-based grammars. %
In \cite{Jurafsky1995scfg}, the authors use grammars to encode semantic rules within limited-vocabulary ASR using a dynamic ad-hoc system. %
More recently, Gandhe et al. \cite{Gandhe2018lmadaptation} estimate n-gram LMs directly from grammars to improve ASR for new application intents in VAs.
Our goal is to integrate the probabilistic grammar within the finite-state transducer (FST) framework due to its ability to efficiently integrate acoustic and language knowledge sources \cite{Aubert2000overview}, where LM and acoustic components are composed dynamically at runtime \cite{Dolfing2001incremental}. The dynamic interpolation of LMs is necessary when different LM components are updated at different cadences. Reducing the size of model updates, by only updating some components of an ASR system, is essential when ASR occurs on-device and bandwidth/disk space is costly. For VAs, the entity query distribution changes faster than the general voice query distribution, and hence, requires more frequent updates. In addition, in order to guarantee efficient decoding, the approaches we consider are limited to those that guarantee determinism (i.e., unlike a dynamicly-generated class-based LM).

Our research questions (RQs) are:
\begin{enumerate*}[label=(\textbf{\footnotesize RQ\arabic*})]%
    \item Can we improve the long tail entity coverage of LMs while operating under model size constraints?
    \item Do size-constrained improvements in entity coverage translate into ASR quality improvements?
\end{enumerate*}
We contribute:
\begin{enumerate*}[label=(\arabic*)]%
    \item analysis on prevalance of entity-oriented queries within VA usage,
    \item a modeling approach---\LongMethodName{} (\emph{``phi RTN''})---that provides a deterministic approximation to a prob. grammar compatible with the FST framework and complementary to n-gram models, and
    \item their experimental evaluation.
\end{enumerate*}

\section{Methodology}
\label{section:methodology}

\newcommand{\Slot}[1]{\${}#1}%
\newcommand{\EntitySlot}{\Slot{entity}}

\newcommand{\Grammar}{G}
\newcommand{\Entities}{E}
\newcommand{\Templates}{T}

\newcommand{\FST}[1]{F_{#1}}

\newcommand{\TokenIdentifier}{w}
\newcommand{\Token}[1]{{\TokenIdentifier{}}_{#1}}

\newcommand{\Vocabulary}{V}

\begin{figure}[t]%
\hfill%
\renewcommand\thesubfigure{\Alph{subfigure}}%
\begin{subfigure}[b]{0.45\columnwidth}%
    \centering\footnotesize%
    \renewcommand{\tabcolsep}{5pt}%
    \resizebox{\textwidth}{!}{%
    \renewcommand{\arraystretch}{0.65}%
    \begin{tabular}{p{5pt}lc}
        \toprule
        \textbf{\#} & \textbf{template} & \textbf{prob.} \\
        \midrule
        (1) & play \EntitySlot{} & $0.4$ \\
        (2) & \EntitySlot{} & $0.2$ \\
        (3) & hey VA \EntitySlot{} & $0.1$ \\
        (4) & hey VA play \EntitySlot{} & $0.1$ \\
        (5) & VA play \EntitySlot{} & $0.1$ \\
        (6) & show me \EntitySlot{} & $0.1$ \\
        \bottomrule
    \end{tabular}}%
    \caption{Template queries.\label{fig:grammar:templates}}%
\end{subfigure}%
\hfill%
\begin{subfigure}[b]{0.45\columnwidth}%
    \centering\footnotesize%
    \renewcommand{\tabcolsep}{6.1985pt}%
    \resizebox{\textwidth}{!}{%
    \renewcommand{\arraystretch}{0.65}%
    \begin{tabular}{p{5pt}lc}
        \toprule
        \textbf{\#} & \textbf{\EntitySlot{}} & \textbf{prob.} \\
        \midrule
        (a) & hip hop rap & $2.7 \cdot 10^{-3}$ \\
        (b) & Adele & $8.0 \cdot 10^{-5}$ \\
        (c) & Drake & $7.9 \cdot 10^{-5}$ \\
        (d) & NBA YoungBoy & $7.4 \cdot 10^{-5}$ \\
        (e) & The Beatles & $6.3 \cdot 10^{-5}$ \\
        \multicolumn{2}{c}{$\cdots$} \\
        (f) & play on Canada & $9.6 \cdot 10^{-9}$ \\
        \bottomrule
    \end{tabular}}%
    \caption{Non-terminal \EntitySlot{}.\label{fig:grammar:entities}}%
\end{subfigure}%
\hfill%
\caption{Grammar of entity-centric media player queries.\label{fig:grammar}}
\end{figure}

We build an entity-centric LM of spoken VA queries that is complementary to a general LM trained on transcribed VA queries. Our goal is to improve the speech recognition of tail entities, while taking into account resource constraints (i.e., model size). ASR LMs assign a non-zero probability $\CondProb{\Token{k}}{\Token{1}, \ldots, \Token{k - 1}}$ to word $\Token{k}$ preceded by left-context $\Token{1}, \ldots, \Token{k - 1}$, with words $\Token{i}$ members of a fixed-size vocabulary $\Vocabulary{}$. During ASR decoding, the LM probability is combined with acoustic information to differentiate between competing hypotheses.
We will now proceed as follows. %
First, we describe how entity-oriented queries can be described as probabilistic grammars (\S\ref{section:methodology:entity_query_grammars}). Next, we provide background on non-deterministic graph grammar representations and their shortcomings (\S\ref{section:methodology:rtn}). Finally, we describe our approach, \LongMethodName{}, that provides an approximation to FST-compatible graph-based grammars (\S\ref{section:methodology:dartn}).

\subsection{Entity-centric Query Grammars}
\label{section:methodology:entity_query_grammars}

An analysis of a month of a random sample anonymized U.S. English query logs from a popular VA shows that over \textbf{\numprint{15}\%} of media player queries that instructed the VA to play a song, album or artist follow the \emph{"play \EntitySlot{}"} template.
Hence, a significant portion of media player queries can be represented using a probabilistic grammar. Estimating a grammar from entity-centric queries extracted from usage logs---that will subsequently be used for LM estimation---has the advantage that it allows us to keep the query templates static, while updating the weighted entity list using external knowledge sources that correlate with usage. In turn, this allows the VA to recognize emerging entities that are not yet included in transcribed training data.
Denote $\Templates{}$ and $\Entities{}$ as the sets of query templates and entities, resp., an example grammar $\Grammar{}$---consisting of templates $\Templates{}$ with slots for entities $\Entities{}$---that we wish to support is depicted in Fig.~\ref{fig:grammar}. The templates in Fig.~\ref{fig:grammar:templates} were extracted from usage logs and subsequently inspected by domain experts, whereas the entity feed in Fig.~\ref{fig:grammar:entities} was extracted from an external knowledge source. We are particularly interested in improving LM coverage for tail entity queries. We define queries whose joint probability falls below the median as tail queries.

\subsection{Recursive Transitive Networks}
\label{section:methodology:rtn}
\newcommand{\RootNonTerminal}{S}
Encoding the grammar depicted in Fig.~\ref{fig:grammar} as a static FST by expanding every occurrence of \emph{\EntitySlot{}} in the templates (Fig.~\ref{fig:grammar:templates}) leads to excessively large models---proportional to the cross-product, $\SetSize{\Templates{}} \cdot \SetSize{\Entities{}}$. Recursive Transitive Networks (RTNs) \cite{Woods1970rtn} offer an alternative representation to encode grammars---and more generally, class-based LMs \cite{Brown1992classlm}---that do not store the cross-product. Instead, an RTN consists of a family of FSTs---each associated with a non-terminal---where non-terminal labels on arcs within the constituent FSTs are recursively replaced by the FST they reference. Non-terminal symbol $\RootNonTerminal{}$ indicates the root non-terminal that belongs to the constituent FST that is explored first. Grammar $\Grammar{}$ (Fig.~\ref{fig:grammar}) can be represented using a RTN with a single level of recursion. %
Unfortunately, RTNs, when used as decoding graphs in a FST decoder, are generally non-deterministic \cite[\S4.5]{Allauzen2012pdt}. An FST is non-deterministic if at a given state there are multiple outgoing arcs that share the same input symbol. Non-determinism occurs within RTNs since there exist multiple paths through the graph that result in the same token sequence. Within the grammar depicted in Fig.~\ref{fig:grammar}, this would be the case within the state that corresponds to context \emph{"hey VA"}---i.e., the wake word contributed by templates $(3)$ and $(4)$---since the next token \emph{"play"} can either match template $(4)$ or entity $(f)$.
Non-determinism is detrimental to efficient ASR decoding as multiple paths can lead to the same hypothesis, and therefore should be avoided.

\subsection{Deterministic Approximate RTNs}
\label{section:methodology:dartn}

\newcommand{\State}{s}
\newcommand{\NextState}{\State{}^\prime}
\newcommand{\OtherState}{\NextState{}}
\newcommand{\ObservedTokens}[1]{\Token{1}, \ldots, \Token{#1}}
\newcommand{\NextToken}[1]{\Token{#1 + 1}}

A key characteristic of entity-oriented VA queries is that carrier phrases (i.e., the templates excluding the non-terminal) are often short and use a limited vocabulary of frequent tokens ($< 100$ unique; see \S\ref{section:experiments:statistics}), whereas the entity names typically consist of ample infrequent tokens ($> 10^4$ unique; see \S\ref{section:experiments:statistics}). We can use this observation to relax the harsh conditions necessary for determinism when using RTNs for modeling entity-oriented queries by imposing a precedence of regular symbols over non-terminal symbols. More specifically, when we are at state $\State{}$ in our RTN after observing $i$ tokens $\ObservedTokens{i}$ and need to transition to the next state $\NextState{}$ after observing the next token $\NextToken{i}$, we will first attempt to match a regular symbol on any arc leaving state $\State{}$, and only explore entering a non-terminal FST, or exiting the current non-terminal FST, if a match cannot be found. Within the FST framework, this behavior can be implemented using $\phi$-transitions \cite{OpenFST2020matchers}, i.e., transitions followed when the requested symbol cannot be matched at the current state, that are typically used to implement back-off n-gram models \cite{Katz1987backoff} using FSTs.

We will now proceed by describing the topology of our model, named \LongMethodName{}, as a two-level RTN, and explain how our model is made deterministic---approximating the actual RTN---by following precedence rules using $\phi$-transitions. In addition, we discuss how we ensure that our model is correctly normalized. Later, at the end of \S\ref{section:results}, we discuss limitations/drawbacks of our approach---and means to alleviate them.

\noindent \textbf{Topology.} %
\newcommand{\TemplateFST}{\FST{\Templates{}}}%
\newcommand{\EntityFST}{\FST{\Entities{}}}%
\newcommand{\UnigramState}{\State{}_{\Prob{\TokenIdentifier{}}}}%
\newcommand{\SmoothingAlpha}{\alpha}%
\newcommand{\SmoothingAlphaComplement}{\overline{\SmoothingAlpha{}}}%
\newcommand{\NumStates}[1]{\Apply{\text{NumStates}}{#1}}%
\newcommand{\OutgoingSymbols}[1]{\Apply{\text{OutgoingSymbols}}{#1}}%
We employ a two-level RTN, where root non-terminal $\RootNonTerminal{}$ is associated with a rigid grammar FST $\TemplateFST{}$ estimated on templates (e.g., Fig.~\ref{fig:grammar:templates}). The transition probabilities at each state are multiplied with discounting factor $\SmoothingAlphaComplement{} = 1 - \SmoothingAlpha{}$ ($0 < \SmoothingAlpha{} < 1$) to allow for out-of-domain utterances. The leftover mass, $\SmoothingAlpha{}$, is distributed over all unseen tokens at the state through the use of a back-off arc that leads to a self-loop unigram state $\UnigramState{}$ that encodes a unigram probability distribution $\Prob{\TokenIdentifier{}}, \forall\, \TokenIdentifier{} \in \Vocabulary{}$. Template FST $\TemplateFST{}$ references non-terminal $\EntitySlot{}$, associated with entity FST $\EntityFST{}$. FST $\EntityFST{}$ is modeled as a regular, non-backoff $n$-gram model over entity names (e.g., Fig.~\ref{fig:grammar:entities}), with probabilities scaled by $\SmoothingAlphaComplement{}$ to allow for out-of-domain entities. The left-over mass at each state in $\EntityFST{}$ will be used for transitioning out of $\EntityFST{}$, back to $\TemplateFST{}$, and possibly $\UnigramState{}$ in the case of out-of-domain entities (see below).

\noindent \textbf{Determinism.} %
We address the RTN determinization issue discussed in \S\ref{section:methodology:rtn} by prioritizing regular symbols over non-terminal entry/exit. More specifically, within $\TemplateFST{}$, when observing non-terminal $\EntitySlot{}$ after context $\ObservedTokens{i}$, with corresponding state $\State{}$, the corresponding sub-FST $\EntityFST{}$ can be reached by following the $\phi$-transition from state $\State{}$. Note that in the absence of $\EntitySlot{}$, the $\phi$-transition leads to the unigram state $\UnigramState{}$. When entering $\EntityFST{}$, we remember at which state we need to return to in $\TemplateFST{}$. Within $\EntityFST{}$, $\phi$-transitions are used to exit sub-FST $\EntityFST{}$ and return to template FST $\TemplateFST{}$. Final states in $\EntityFST{}$ are removed, and instead, at each state $\State{}$ in $\EntityFST{}$, the FST state's final probability mass is combined with the leftover mass $\SmoothingAlpha{}$ and distributed over regular symbols not observed at $\State{}$.

\noindent \textbf{Normalization.} %
To ensure that the resulting model is properly normalized, i.e. $\sum_{\TokenIdentifier{} \in \Vocabulary} \CondProb{\TokenIdentifier{}}{\State{}} = 1$ for all states $\State{}$, we follow a strategy similar to back-off n-gram models \cite{Katz1987backoff}. More precisely, we assign a weight to the $\phi$-transition leaving state $\State{}$ such that the probability mass assigned to unobserved events at $\State{}$, when following $\phi$-transitions recursively, are scaled to fit into the leftover mass at state $\State{}$.
For $\phi$-transitions leading from $\TemplateFST{}$ to $\EntityFST{}$, the weights can be computed statically when the model is constructed---since we know the start state in $\EntityFST{}$ the $\phi$-transition leads to, and the state from $\EntityFST{}$ that leads back to $\TemplateFST{}$ after following $\phi$ once more.
However, computation of the weights for $\phi$-transitions that exit from $\EntityFST{}$ to $\TemplateFST{}$ (excl. start state in $\EntityFST{}$) is more difficult---due to the dependency between the state in $\EntityFST{}$ from which we are exiting and the state in $\TemplateFST{}$ we are returning to. If we were to compute these $\phi$-transition weights statically, we would need to store $\BigOh{\NumStates{\TemplateFST{}} \cdot \NumStates{\EntityFST{}}}$ weights. To avoid the prohibitively expensive storage costs, we compute the weights of $\phi$-transitions leading from $\EntityFST{}$ back to $\TemplateFST{}$ partially at runtime. Partial computation of the weights is achieved by pre-computing the marginal unigram probability, $\sum_{\TokenIdentifier{} \in \OutgoingSymbols{\State{}}} {\Prob{\TokenIdentifier{}}}$, for all explicit symbols defined at each state $\State{}$ in $\EntityFST{}$ when the model is constructed. At runtime, since we know the leftover mass at state $\State{}$ in $\EntityFST{}$ equals $\SmoothingAlpha{} + \CondProb{\text{final}}{\State{}}$, we thus only need to consider explicit events defined at state $\NextState{}$ in $\TemplateFST{}$ that the $\phi$-transition leaving $\EntityFST{}$ leads to, in order to update the partial weight. Since in our setting, $\TemplateFST{}$ is sparse, the computation involved in this operation is negligible.
Note that \LongMethodName{} does not support templates with two or more consecutive non-terminals since that would lead to significantly more computation.

\section{Experimental setup}
\label{section:experiments}

\newcommand{\VectorFST}{\compacttexttt{vector}}
\newcommand{\CompactAcceptorFST}{\compacttexttt{compact}}
\newcommand{\NGramFST}{\compacttexttt{ngram}}
\newcommand{\PhiRTN}{\compacttexttt{phi-rtn}}

\newcommand{\ArpaToFst}{\compacttexttt{arpa2fst}}

\newcommand{\NumTemplates}{293}
\newcommand{\NumEntities}{2608460}
\newcommand{\NumUniqueTemplateTokens}{77}
\newcommand{\NumUniqueEntityTokens}{230321}

\newcommand{\Head}{head}
\newcommand{\Torso}{torso}
\newcommand{\Tail}{tail}

\newcommand{\TestSampleSize}{10k}

\subsection{Entity-centric Query Grammar}
\label{section:experiments:statistics}

We focus on media player-type queries, where users instruct a VA to interact with audio content (e.g., songs, playlists).  %
Our media player query grammar consists of a weighted list of \numprint{\NumTemplates{}} templates (\numprint{\NumUniqueTemplateTokens{}} unique tokens) that reference an entity slot and a weighted list of \numprint{\NumEntities{}} entities (\numprint{\NumUniqueEntityTokens{}} unique tokens) that can fill the slot.
The templates were derived from high-frequency use-cases in a representative sample of query logs by domain experts with prior probabilities proportional to their presence in user requests. %
As shown in \S\ref{section:methodology:entity_query_grammars}, a small number of templates can model a significant proportion of user queries. The list of media entities was extracted from the catalog of a popular media service, with probabilities based on interactions.

\subsection{Evaluation Sets}
\label{section:experiments:evaluation}

For evaluation, we partition the set of expanded synthetic queries according to the rank percentiles of their joint probabilities (i.e., $\Prob{\text{template}} \cdot \Prob{\text{entity}}$): \Head{} (top-10\%), \Torso{} (between top-10\% and the median), and \Tail{} (bottom-50\%). Subsequently we sample \TestSampleSize{} queries from each stratum for %
\begin{enumerate*}[label=(\arabic*)]
    \item evaluating the trade-off between perplexity and query LM size across the compared approaches (see below), and
    \item word error rate (WER) evaluation where we linearly interpolate the query LMs with our main ASR LM, with weight coefficients $0.05$ and $0.95$, resp., on audio generated from the randomly sampled queries with a Neural Text-To-Speech system described in \cite{Sivanand2021ondeviceTTS} to measure the effectiveness of our approach. %
    The 95/5\% weight distribution was chosen to allow the query LM to influence recognition while not dominating it. Within experiments not included in this paper, we verified that a 5\% weight is sufficient for any model to maximize its impact without negatively impacting out-of-domain utterances.
\end{enumerate*}
In addition, we also evaluate WER using the same setup on a uniform random sample of VA queries to ensure there is no regression.
For (hyper-)parameter optimization, we build a development set by first sampling \TestSampleSize{} queries from each stratum---that do not overlap with the test set---and taking the union.
When interpolating multiple query LMs with the main ASR LM, the $0.05$ weight is divided across the query LMs by maximizing likelihood on the dev. set using L-BFGS-B \cite{Zhu1997lbfgsb}.

\subsection{Approaches under comparison}
\label{section:experiments:fst_representations}
\label{section:experiments:system}

We compare the following approaches that provide deterministic decoding graphs, where we sweep over hyperparameters that influence model size, for the task of recognition of spoken entity-centric VA queries:
\begin{enumerate*}[label=(\arabic*)]
\item We construct \LongMethodName{} models directly from the grammar of templates and entities ($n = 2, 3, 4$).
\item We train back-off n-gram LMs ($n = 2, 3, 4$) using SRILM with Witten-Bell smoothing, and subsequently perform entropy pruning \cite{Stolcke1998pruning} (with threshold $\theta \in \{ 4^{-i} \mid 4 \leq i < 20\} \cup \{0\}$) to reduce the model size, on the full set of expanded synthetic queries (\S\ref{section:introduction}, \S\ref{section:methodology:entity_query_grammars}). Witten-Bell smoothing was chosen since it puts less reliance on the absolute counts than other smoothing methods (e.g., Good-Turing) and is therefore more suitable for synthetic data. The grammar-generated phrases are weighted according to pseudo-counts that were obtained by rescaling their probabilities such that the lowest pseudo-count equals \numprint{1}. To avoid the filtering of n-grams with low pseudo-counts, we disabled the n-gram discounting cutoffs (i.e., $\text{gt}n\text{min} = 1 \, \forall \, n$).
After n-gram model generation, we convert the obtained ARPA model to OpenFST \cite{Allauzen2007openfst} models using Kaldi's \ArpaToFst{} tool \cite{Povey2011kaldi}. To evaluate the space requirement of the model, we compare multiple FST formats: %
\begin{enumerate*}[label=(\alph*)]
    \item free-form formats, such as the mutable \VectorFST{} and the immutable space-optimized \CompactAcceptorFST{} acceptor formats,
    \item the specialized \NGramFST{} format \cite{Sorensen2011unary} that uses the LOUDS encoding to efficiently represent n-gram back-off models.
\end{enumerate*}
For the free-form formats (\VectorFST{}, \CompactAcceptorFST{}), we apply FST minimization after \ArpaToFst{} in order to further reduce FST size.
For generation of \NGramFST{} FSTs, we skip the removal of redundant states during \ArpaToFst{}, such that the intermediate output FST retains the necessary information about back-off, and finally use the \compacttexttt{fstconvert} utility to obtain a \NGramFST{} FST.
Note that all FSTs generated from the same n-gram model will be qualitatively identical and only differ in disk space requirements.
\end{enumerate*}
$\;$ \textbf{ASR system.} We used a CNN acoustic model \cite{Huang2020sndcnn}, a 4-gram LM with Good-Turing smoothing in the 1st pass, and the same LM---in addition to the query LMs---interpolated with a Neural LM in the 2nd pass using a dynamic weighting scheme.
To build the 4-gram LM, component 4-gram models built from data sources ($>$ 10B manually/auto. transcribed anonymized VA requests) were combined \cite{Pusateri2019interpolation} on held out manually transcribed VA data \cite{Bacchiani2003unsupervised}.

\section{Results \& Discussion}
\label{section:results}

\newcommand{\RQRef}[1]{\textbf{\footnotesize RQ#1}}

We answer the research questions asked in \S\ref{section:introduction} as follows.

%
%

\begin{figure*}[ht]
    \centering
    \begin{subfigure}[b]{0.32\textwidth}
        \centering
        \includegraphics[width=\textwidth]{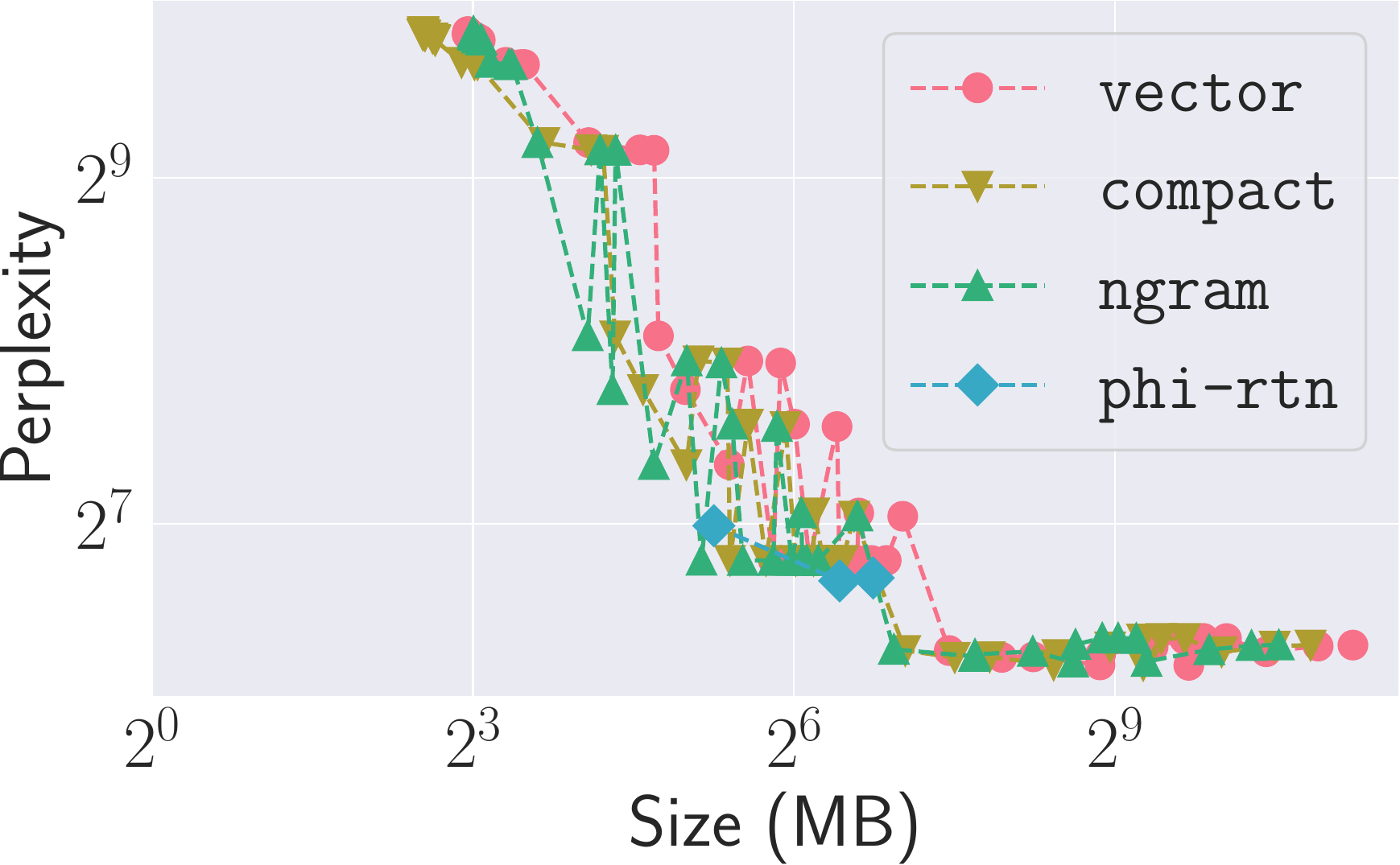}
        \caption{Head (top-\numprint{10}\%)}
        \label{fig:tradeoff:head}
    \end{subfigure}
    \hfill
    \begin{subfigure}[b]{0.32\textwidth}
        \centering
        \includegraphics[width=\textwidth]{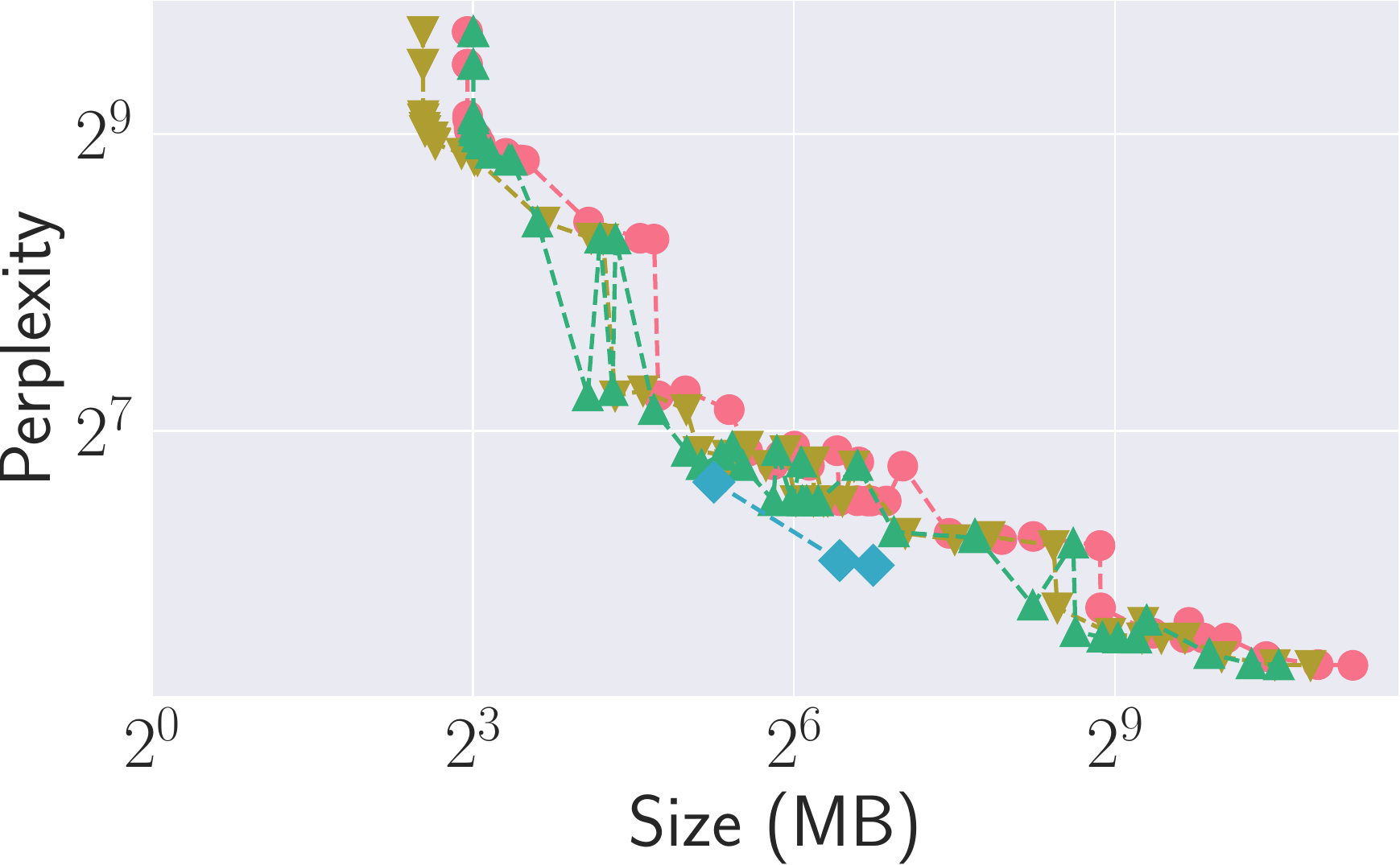}
        \caption{Torso (between top-\numprint{10}\% and \numprint{50}\%)}
        \label{fig:tradeoff:torso}
    \end{subfigure}
    \hfill
    \begin{subfigure}[b]{0.32\textwidth}
        \centering
        \includegraphics[width=\textwidth]{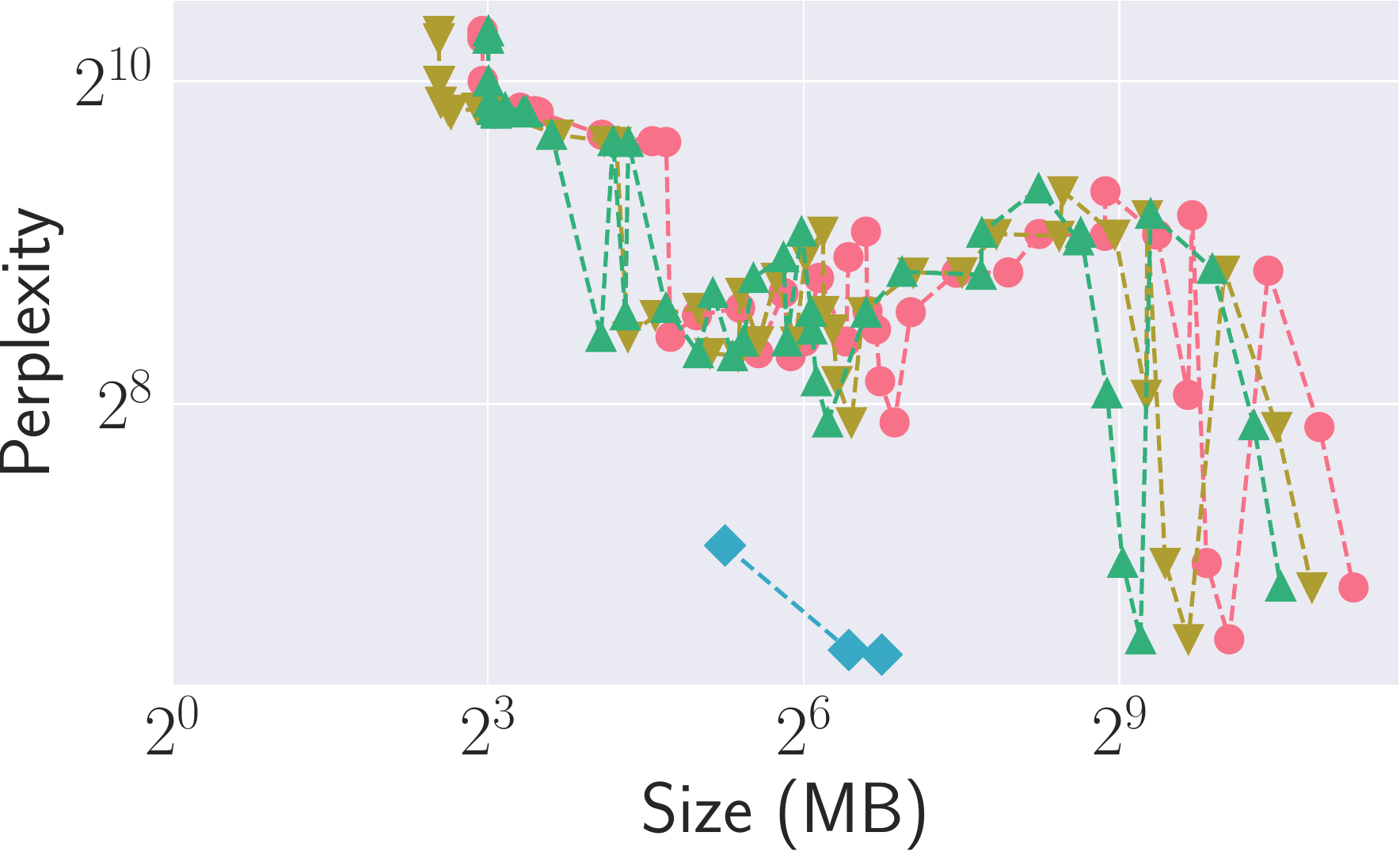}
        \caption{Tail (bottom-\numprint{50}\%)}
        \label{fig:tradeoff:tail}
    \end{subfigure}
    \caption{Model size vs. perplexity trade-off on dev. sets (\S\ref{section:experiments:evaluation}) (closer to bottom-left is better) as we sweep over hyper-params (\S\ref{section:experiments:fst_representations}).\label{fig:tradeoff}}
\end{figure*}

\noindent%
\RQRef{1}: %
Fig.~\ref{fig:tradeoff} shows the trade-off between perplexity and model size for the various approaches (\S\ref{section:experiments:fst_representations}) used to encode back-off LMs (\VectorFST{}, \CompactAcceptorFST{}, \NGramFST{}), and our method, \LongMethodName{} (\PhiRTN{}) on head, torso and tail dev. sets (\S\ref{section:experiments:evaluation}). When considering the back-off LMs for each dev. set, regardless of FST format, the trade-off curves all follow similar shapes and only differ by a near-constant offset on the x-axis. %
For the back-off LMs, the perplexity/size curve oscillates since we sweep over both n-gram order and pruning thresholds---combined within a single curve per approach.
On the head of the distribution (Fig.~\ref{fig:tradeoff:head}), back-off LMs quickly reach low perplexity at small model sizes ($\smallsim2^{7}\text{MB}$). However, when we consider the torso/tail (Fig.~\ref{fig:tradeoff:torso},~\ref{fig:tradeoff:tail}), the curve becomes less sharp and only converges to the minimum when using large models ($\smallsim2^{10}\text{MB}$) with nearly no pruning. %
Across all back-off LM FST formats, the \NGramFST{} format is most economical, followed by \CompactAcceptorFST{} and then \VectorFST{}.
Our method, \PhiRTN{}, provides approximately the same coverage as \NGramFST{} on the head set (Fig.~\ref{fig:tradeoff:head}) at $\smallsim2^{6}\text{MB}$. However, on the tail set (Fig.~\ref{fig:tradeoff:tail}), \PhiRTN{} obtains about an order of magnitude lower perplexity, and consequently, better coverage, at a model size of $\smallsim2^{6}\text{MB}$ compared to the back-off LMs.
The lower perplexity of back-off LMs on the head set, and their higher perplexity on the tail, is due to the following dual-effect: %
\begin{enumerate*}[label=(\alph*)]
    \item back-off smoothing methods rely on absolute counts to evaluate the reliability of empirical prob. estimates, and hence, are generally biased towards high-prob. events, and
    \item entropy pruning favors the removal of low-prob. events.
\end{enumerate*}
Even in the case that we were to modify the smoothing/pruning methods to be less favorable w.r.t. the head of the distribution, the size of back-off LMs is proportional to the number of explicitly-defined probs., and hence, the curves in Fig.~\ref{fig:tradeoff:tail} would not change significantly.
We answer our first RQ as follows: while back-off LMs are unsuitable to improve long tail entity coverage when operating under model size constraints, \LongMethodName{} provides significantly better coverage. Since the worst-case model size of \LongMethodName{} is asymptotically bounded by the number of unique entities, we can choose the entity set to be included according to the target model size.

%
%

\newcommand{\EnumMethodName}{{\footnotesize \PhiRTN{}}}
\newcommand{\NGramMethodName}{{\footnotesize \NGramFST{}}}
\newcommand{\SmallNGramMethodName}{{\footnotesize \NGramMethodName{}$^{(<)}$}}
\newcommand{\MediumNGramMethodName}{{\footnotesize \NGramMethodName{}$^{(\smallsim)}$}}
\newcommand{\LargeNGramMethodName}{{\footnotesize \NGramMethodName{}$^{(>)}$}}

\begin{table}[ht]
    \caption{Overview of methods under comparison for WER evaluation, how they were selected, and their hyper-parameters.\label{tbl:asr_methods}}%
    \centering%
    \footnotesize%
    \renewcommand{\arraystretch}{0.65}%
    \setlength\tabcolsep{3pt}%
    \begin{tabular}{lp{3.5cm}l}%
    \toprule%
    \textbf{Name} & \textbf{Selection criterion} & \textbf{Hyper-params} \\
    \midrule%
    \textbf{\scriptsize \EnumMethodName{}} & Based on Fig.~\ref{fig:tradeoff} & $n = 3$ \\
    \textbf{\scriptsize \SmallNGramMethodName{}} & $\smallsim{}\frac{1}{2}\times$ size \textbf{\EnumMethodName{}} & $n = 3$, $\theta = 6\text{e-}{8}$ \\
    \textbf{\scriptsize \MediumNGramMethodName{}} & $\smallsim{}$ size \textbf{\EnumMethodName{}} & $n = 4$, $\theta = 1\text{e-}{8}$ \\
    \textbf{\scriptsize \LargeNGramMethodName{}} & $\smallsim{}2\times$ size \textbf{\EnumMethodName{}} & $n = 4$, $\theta = 4\text{e-}{9}$ \\
    \bottomrule%
    \end{tabular}
\end{table}
\vspace*{-1.5em}%
\begin{table}[ht]
    \caption{Error-rates using different models (Table~\ref{tbl:asr_methods}) of media player queries, interpolated with the main LM at runtime. Letters between brackets denote sub-sections.\label{tbl:asr_accuracy}}%
    \centering%
    \newcommand{\SubTableLabel}[2]{\speciallabel{%
        #2}{tbl:asr_accuracy:#1}\textbf{[#2]}}%
    \newcommand{\MethodBaselineTitle}{\textbf{Main LM only}}%
    \newcommand{\MethodSmallNGramTitle}{$\;\;\;$\textbf{\scriptsize \SmallNGramMethodName{}}}%
    \newcommand{\MethodMediumNGramTitle}{$\;\;\;$\textbf{\scriptsize \MediumNGramMethodName{}}}%
    \newcommand{\MethodLargeNGramTitle}{$\;\;\;$\textbf{\scriptsize \LargeNGramMethodName}}%
    \newcommand{\MethodEnumTitle}{$\;\;\;$\textbf{\scriptsize \EnumMethodName{}}}%
    \newcommand{\MethodEnumSmallNGramCombinationTitle}{$\;\;\;$\textbf{\scriptsize \EnumMethodName{} + \SmallNGramMethodName{}}}%
    \newcommand{\MethodEnumMediumNGramCombinationTitle}{$\;\;\;$\textbf{\scriptsize \EnumMethodName{} + \MediumNGramMethodName{}}}%
    \newcommand{\MethodEnumLargeNGramCombinationTitle}{$\;\;\;$\textbf{\scriptsize \EnumMethodName{} + \LargeNGramMethodName{}}}%
    \newcommand{\SectionEnumOnlyTitle}{\textit{+ \LongMethodName{} only} \SubTableLabel{enum}{2a}}%
    \newcommand{\SectionSmallBudgetTitle}{\textit{+ small N-Gram/\LongMethodName{} combinations} \SubTableLabel{small}{2b}}%
    \newcommand{\SectionMediumBudgetTitle}{\textit{+ medium N-Gram/\LongMethodName{} combinations} \SubTableLabel{medium}{2c}}%
    \newcommand{\SectionLargeBudgetTitle}{\textit{+ large N-Gram/\LongMethodName{} combinations} \SubTableLabel{large}{2d}}%
    \renewcommand{\arraystretch}{0.55}%
    \setlength\tabcolsep{3pt}%
    \footnotesize%
    \resizebox{0.95\columnwidth}{!}{%
    \begin{tabular}{@{}lrrrrrrr@{}}%
\toprule%
\multirow{2}{*}{\textbf{Method}}&\textbf{Size}&\multicolumn{2}{c}{\textbf{Head}}&\multicolumn{2}{c}{\textbf{Torso}}&\multicolumn{2}{c}{\textbf{Tail}}\\%
&\textbf{\scriptsize (MB)}&\textbf{\scriptsize WER}&\textbf{\scriptsize SER}&\textbf{\scriptsize WER}&\textbf{\scriptsize SER}&\textbf{\scriptsize WER}&\textbf{\scriptsize SER}\\%
\midrule%
\MethodBaselineTitle{}&{-}&10.7&38.2&12.1&47.2&13.2&51.6\\%
\midrule%
\multicolumn{8}{@{}l}{\SectionEnumOnlyTitle{}}\\%
\midrule%
\MethodEnumTitle{}&86.1&5.3&18.2&5.7&20.8&6.3&23.2\\%
\midrule%
\multicolumn{8}{@{}l}{\SectionSmallBudgetTitle{}}\\%
\midrule%
\MethodSmallNGramTitle{}&42.9&5.2&18.8&6.0&22.6&7.0&27.2\\%
\MethodEnumSmallNGramCombinationTitle{}&129.1&5.0&17.6&5.6&20.7&6.1&23.0\\%
\midrule%
\multicolumn{8}{@{}l}{\SectionMediumBudgetTitle{}}\\%
\midrule%
\MethodMediumNGramTitle{}&96.6&5.0&17.7&5.8&22.1&6.8&26.0\\%
\MethodEnumMediumNGramCombinationTitle{}&182.8&5.0&17.2&5.5&20.5&6.0&22.7\\%
\midrule%
\multicolumn{8}{@{}l}{\SectionLargeBudgetTitle{}}\\%
\midrule%
\MethodLargeNGramTitle{}&205.7&4.9&17.0&5.7&21.4&6.7&26.1\\%
\MethodEnumLargeNGramCombinationTitle{}&291.8&4.9&16.9&5.5&20.3&6.0&22.6\\\bottomrule%
\end{tabular}%
    }%
\end{table}

\noindent%
\RQRef{2}: %
We now investigate whether the increase in tail coverage provided by \LongMethodName{} translates into improved ASR performance. Following the observations made from Fig.~\ref{fig:tradeoff} on the dev. sets, we choose the \PhiRTN{} model with $n = 3$ as it provides a good coverage/size trade-off. We choose \NGramFST{} models relative to the size of the \PhiRTN{} model by taking the \NGramFST{} model closest to the target size (e.g., half size of \PhiRTN{}) and pick the \NGramFST{} within a $\epsilon = 5\text{MB}$ window with lowest perplexity on the dev. set (\S\ref{section:experiments:evaluation}).
Table~\ref{tbl:asr_methods} shows the selected models and selection criteria/hyperparameters. In Table~\ref{tbl:asr_accuracy}, we evaluate the \LongMethodName{} and back-off LMs interpolated individually with the main ASR LM, and combinations thereof (\S\ref{section:experiments:evaluation}). Including \PhiRTN{} results in a significant WER improvement across all test sets, when compared to the main LM alone. This is unsurprising, since no media player-specific data was added to the main LM---except the queries occurring in transcribed data (end of \S\ref{section:experiments:system})---as the models are complementary (\S\ref{section:introduction}). When comparing \PhiRTN{} and same-sized \MediumNGramMethodName{} (Table~\ref{tbl:asr_accuracy:medium}), we note that while \PhiRTN{} performs signficantly better on the tail, \MediumNGramMethodName{} wins on the head. Can a combination of both models result in overall better performance on all test sets? Comparing \LargeNGramMethodName{} (Table~\ref{tbl:asr_accuracy:large}) with \PhiRTN{} + \MediumNGramMethodName{} (Table~\ref{tbl:asr_accuracy:medium}), the \numprint{11}\% smaller combination performs similarly on the head test set, but achieves a 10\% relative WER reduction on the tail set.
Finally, we did not observe a regression on the uniform sample of VA queries (\S\ref{section:experiments:statistics}) in terms of WER and end-to-end latency.
We answer our 2nd RQ: improvements in model coverage translate to lower WER. In addition, the sum is greater than the parts: %
combining both methods yields an ASR quality increase at less space than would be obtained when using larger back-off LMs.

\noindent \textbf{Discussion.} %
Since \LongMethodName{} stores templates and entities as two separate sub-graphs, and $\SetSize{\Entities{}} \gg \SetSize{\Templates{}}$, the worst-case model size of \LongMethodName{} is asymptotically bounded by the number of unique entities.
The space-saving of \LongMethodName{} over back-off LMs (\S\ref{section:experiments:fst_representations}) occurs since back-off LMs need to explicitly represent the relation between template contexts and entities (i.e., explicit probabilities).
\LongMethodName{} is suitable when the language used in the templates is sufficiently limited. Within our experiments, the \LongMethodName{} model provides coverage for 99\% of queries represented by the grammar (\S\ref{section:experiments:statistics}). However, in other applications where template language may be more complex, \LongMethodName{} may not be the best fit.
Even though \LongMethodName{} has limitations, its inclusion in VA ASR systems allows the recognition of long tail entities that would otherwise not be recognized at all.
Finally, collisions between the template/entity vocabulary that would lead to reduced coverage can be detected at training time and addressed by, e.g., rewriting the template/entity lists or including the affected query in the training data of a complimentary n-gram LM.

\section{Conclusions}
\label{section:conclusions}

We introduced \LongMethodName{}---a grammar-backed LM with low disk space cost that integrates with FST decoders and improves coverage for long tail entities in VA ASR scenarios. \LongMethodName{} avoids the explicit expansion of non-terminals at creation time and stores a single sub-graph per non-terminal, leading to a low disk footprint. %
Despite limitations (end of \S\ref{section:results}), we show that \LongMethodName{} is complementary to LMs trained on the expanded grammar---allowing us to reap the benefits of both models---and improves WER on long tail entity queries by 10\%.
Future work includes the expansion of \LongMethodName{} to non-English languages where grammatical agreements and morphology may require specialized solutions.
$\;$ \textbf{Acknowledgements} We thank Amr Mousa, Barry Theobald, Man-Hung Siu, and the anonymous reviewers for their comments and feedback.

\bibliographystyle{IEEEtran}

\bibliography{interspeech2022-phi_rtn}

\end{document}